\documentclass[5p,twocolumn]{elsarticle}
\pdfoutput=1
\usepackage[table,xcdraw]{xcolor}
\usepackage{lineno}
\usepackage{textcomp}
\usepackage[utf8]{inputenc}
\usepackage[T1]{fontenc}
\usepackage{url}
\usepackage{parskip} 
\usepackage{booktabs}
\usepackage{tikz,pgfplots}
\usepackage{color}
\usepackage{verbatim}
\usepackage{multirow}
\usepackage{lipsum}
\usepackage{subcaption}
\usepackage{booktabs}
\usepackage{afterpage}

\modulolinenumbers[20]

\begin{document}
\begin{frontmatter}

\title{Deep Multi-scale Location-aware 3D Convolutional Neural Networks for Automated Detection of Lacunes of Presumed Vascular Origin}

\author[uni,hospital]{Mohsen Ghafoorian\footnotemark[1]}
\author[hospital]{Nico Karssemeijer}
\author[uni]{Tom Heskes}
\author[neuro]{Mayra Bergkamp}
\author[neuro]{Joost Wissink}
\author[hospital]{Jiri Obels}
\author[neuro]{Karlijn Keizer}
\author[neuro]{Frank-Erik de Leeuw}
\author[hospital]{Bram van Ginneken}
\author[uni]{Elena Marchiori}
\author[hospital]{Bram Platel}

\address[uni]{Institute for Computing and Information Sciences, Radboud University, Nijmegen, The Netherlands}
\address[hospital]{Diagnostic Image Analysis Group, Department of Radiology and Nuclear Medicine, Radboud University Medical Center, Nijmegen, The Netherlands}
\address[neuro]{Donders Institute for Brain, Cognition and Behaviour, Department of Neurology, Radboud University Medical Center, Nijmegen, The Netherlands}
\begin{abstract}
Lacunes of presumed vascular origin (lacunes) are associated with an increased risk of stroke, gait impairment, and dementia and are a primary imaging feature of the small vessel disease. Quantification of lacunes may be of great importance to elucidate the mechanisms behind neuro-degenerative disorders and is recommended as part of study standards for small vessel disease research. However, due to the different appearance of lacunes in various brain regions and the existence of other similar-looking structures, such as perivascular spaces, manual annotation is a difficult, elaborative and subjective task, which can potentially be greatly improved by reliable and consistent computer-aided detection (CAD) routines.

In this paper, we propose an automated two-stage method using deep convolutional neural networks (CNN). We show that this method has good performance and can considerably benefit readers. We first use a fully convolutional neural network to detect initial candidates. In the second step, we employ a 3D CNN as a false positive reduction tool. As the location
information is important to the analysis of candidate structures, we further equip the network with contextual information using multi-scale analysis and integration of explicit location features. We trained, validated and tested our networks on a large dataset of 1075 cases obtained from two different studies. Subsequently, we conducted an observer study with four trained observers and compared our method with them using a free-response operating characteristic analysis. Shown on a test set of 111 cases, the resulting CAD system exhibits performance similar to the trained human observers and achieves a sensitivity of 0.974 with 0.13 false positives per slice. A feasibility study also showed that a trained human observer would considerably benefit once aided by the CAD system.
\end{abstract}


\end{frontmatter}
 \biboptions{sort&compress}
\section{Introduction}
\footnotetext[1]{Email: mohsen.ghafoorian@radboudumc.nl, phone: +31 243655793, fax: +31 24 3652289}
Lacunes of presumed vascular origin (lacunes), also referred to as lacunar strokes or silent brain infarcts, are frequent imaging features on scans of elderly patients and are associated with an increased risk of stroke, gait impairment, and dementia \cite{vermeer2003silent, santos2009differential, choi2012silent, snowdon1997brain}. Lacunes are presumed to be caused by either symptomatic or silent small subcortical infarcts, or by small deep haemorrhages \cite{franke1991residual} and together with white matter hyperintensities, microbleeds, perivascular spaces and brain atrophy are known to be imaging biomarkers that signify the small vessel disease (SVD) \cite{wardlaw2008lacune}.\\
Lacunes are defined as round or ovoid subcortical fluid-filled cavities of between 3 mm and about 15 mm in diameter with signal intensities similar to cerebrospinal fluid (CSF) \cite{wardlaw2013neuroimaging}. On fluid-attenuated inversion recovery (FLAIR) images, lacunes are mostly represented by a central CSF-like hypointensity with a surrounding hyperintense rim; although the rim may not always be present \cite{wardlaw2013neuroimaging}. In some cases, the central cavity is not suppressed on the FLAIR image and hence the lesion might appear entirely hyperintense, while a clear CSF-like intensity appears on other sequences such as T1-weighted or T2-weighted MR images \cite{moreau2012cavitation}. \\
Wardlaw et al. \cite{wardlaw2013neuroimaging} propose measurements of the number and location of lacunes of presumed vascular origin as part of analysis standards for neuroimaging features of SVD studies. However, this is known to be a challenging highly subjective task since the lacunes can be difficult to differentiate from the perivascular spaces, another SVD imaging feature. Perivascular spaces are also areas filled by cerebrospinal fluid, that even though they are often smaller than 3 mm, they could enlarge up to 10 to 20 mm \cite{wardlaw2013neuroimaging}. Although perivascular spaces naturally lack the hyperintense rim, such a rim could also surround perivascular spaces when they pass through an area of white matter hyperintensity \cite{awad1986incidental}. \\
Considering the importance, difficulty and hence potential subjectivity of the lacune detection task, assistance from a computer-aided detection (CAD) system may increase overall user performance. Therefore, a number of automated methods have been proposed:\\
Yokoyama et al. \cite{yokoyama2007development} developed two separate methods for detection of isolated lacunes and lacunes adjacent to the ventricles, using threshold-based multiphase binarization and a top-hat transform respectively. Later on, Uchiyama et al. employed false positive reducers on top of the previously mentioned method, describing each candidate with 12 features accompanied with a rule-based and a support vector machine classifier \cite{uchiyama2007computer} or alternatively a rule-based and a three-layered neural network followed by an extra modular classifier \cite{uchiyama2007improvement}. In another study Uchiyama et al. used six features and a neural network for discriminating lacunes from perivascular spaces \cite{uchiyama2009cad,uchiyama2008computer}. They also showed that the performance of radiologists without a CAD system could be improved once the CAD system detections were exposed to the radiologists \cite{uchiyama2012computer}. Another false positive reduction method using template matching in the eigenspace was recently utilized by the same group \cite{uchiyama2015eigenspace}. Finally, Wang et al. \cite{wang2012multi} detect lacunes by dilating the white matter mask and using a rule-based pruning of false positives considering their intensity levels compared to the surrounding white matter tissue.\\\\
Deep neural networks \citep{lecun2015deep, schmidhuber2015deep} are biologically inspired learning structures and have so far claimed human level or super-human performances in several different domains \citep{cirecsan2012multi, he2015delving, taigman2014deepface, ciresan2012deep, cirecsan2013multi}. Recently deep architectures and in particular convolutional neural networks (CNN) \citep{lecun1998gradient} have attracted enormous attention also in the medical image analysis field, given their exceptional ability to learn discriminative representations for a large variety of tasks. Therefore a recent wave of deep learning based methods has appeared in various domains of medical image analysis \cite{greenspan2016guest}, including neuro-imaging tasks such as brain extraction \citep{kleesiek2016deep}, tissue segmentation \citep{zhang2015deep,moeskops2016automatic}, tumor segmentation \citep{pereira2016brain,havaei2016brain}, microbleed detection \cite{dou2016automatic} and brain lesion segmentation \citep{ghafoorian2016location,brosch2016deep,brosch2015deep,kamnitsas2016efficient,ghafoorian2016non}.\\\\
In this paper, we propose a two-stage application of deep convolutional networks for the detection of lacunes. We use a fully convolutional network \cite{long2015fully} for candidate detection and a 3D convolutional network for false positive reduction. Since the anatomical location of imaging features is of importance in neuro-image analysis (e.g. for the detection of WMHs \cite{ghafoorian2015small}), we equip the CNN with more contextual information by performing multi-scale analysis as well as adding explicit location information to the network. To evaluate the performance of our proposed method and compare it to trained human observers, we perform an observer study on a large test set of 111 subjects with different underlying disorders. 

\section{Materials}
Data for training and evaluation of our method comes from two different studies: the Radboud University Nijmegen Diffusion tensor and Magnetic resonance imaging Cohort (RUNDMC) and the Follow-Up of transient ischemic attack and stroke patients and Unelucidated Risk factor Evaluation study (FUTURE). The RUNDMC \cite{Norden11c} investigates the risk factors and clinical consequences of SVD in individuals 50 to 85 years old without dementia and the FUTURE \cite{rutten2011risk} is a single-centre cohort study on risk factors and prognosis of young patients with either transient ischemic attack, ischemic stroke or hemorrhagic stroke. We collected 654 and 421 MR images from the RUNDMC and the FUTURE studies respectively, summing up to 1075 scans in total.
\subsection{Magnetic Resonance Imaging}
For each subject we used a 3D T1 magnetization-prepared rapid gradient-echo (MPRAGE) with voxel size of 1.0$\times$1.0$\times$1.0 mm and a FLAIR pulse sequence
with voxel size 1.0$\times$1.2$\times$3.0 mm (including a slice gap of 0.5 mm).

\begin{figure*}[t]
\centering
\centerline
{
	\includegraphics[width=7in]{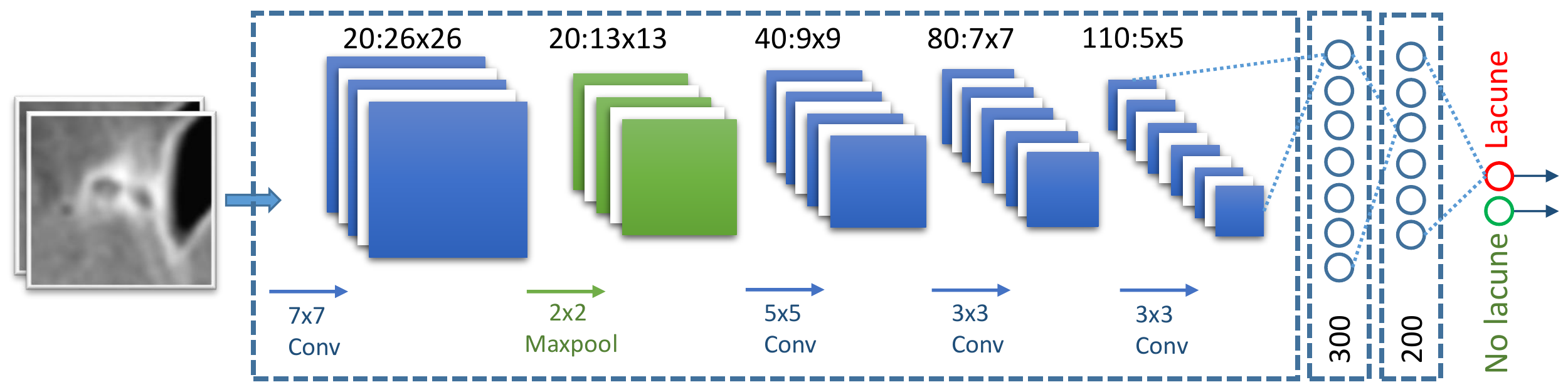}
}
\caption{CNN architecture for candidate detection.}
\label{fig:candDetArch}
\end{figure*}

\subsection{Training, Validation and Test Sets}
We randomly split the total 1075 cases into three sets of size 868, 96 and 111 scans for training, validation and test purposes respectively.

\subsection {Reference Annotations}
Lacunes were delineated for all the images in the training and validation sets in a slice by slice manner by two trained raters (one for the RUNDMC and another for the FUTURE dataset), following the definitions provided in the SVD neuro-imaging study standards \cite{wardlaw2013neuroimaging}.

\subsection{Preprocessing}
We performed the following pre-processing steps before supplying the data to our networks.

\subsubsection{Image Registration}
Due to possible movement of patients during scanning, the image coordinates of the T1 and FLAIR modalities might not represent the same location. Thus we performed a rigid registration of T1 to FLAIR image for each subject, by optimizing the mutual information with trilinear interpolation resampling. For this purpose, we used FSL-FLIRT \citep{jenkinson2001global}. Also to obtain a mapping between patient space and an atlas space, all subjects were non-linearly registered to the ICBM152 atlas \citep{mazziotta2001four} using FSL-FNIRT \citep{jenkinson2012fsl}. 

\subsubsection{Brain Extraction}
To extract the brain and exclude other structures, such as skull, eyes, etc., we applied FSL-BET \citep{smith2002fast} on T1 images. The resulting masks were then transformed using registration transformations and were applied to the FLAIR images.

\subsubsection{Bias Field Correction}
We applied FSL-FAST \citep{zhang2001segmentation}, which uses a hidden Markov random field and an associated expectation-maximization algorithm to correct for spatial intensity variations caused by RF inhomogeneities.

\section{Methods}
Our proposed CAD scheme consists of two phases, a candidate detector and a false positive reducer, for both of which, we employ convolutional neural networks. The details for each subproblem are expanded in the following subsections.

\subsection{Candidate Detection}
As a suitable candidate detector, a method should be fast, highly sensitive to lacunes, while keeping the number of candidates relatively low. To achieve these, we formulated the candidate detection as a segmentation problem and used a CNN for this segmentation task. A CNN would likely satisfy all the three criteria above: CNNs have shown to be great tools for learning discriminative representation of the input pattern. Additionally, once CNNs are formulated in a fully convolutional form \cite{long2015fully}, they can also be very fast in providing dense predictions for image segmentation (in order of a few seconds for typical brain images).

\subsubsection{Sampling}
We captured 51$\times$51 patches to describe a local neighborhood of each voxel we took as a sample, from both the FLAIR and T1 images. As positive samples, we picked all the voxels in the lacune masks and augmented them by flipping the patch horizontally. We randomly sampled negative patches within the brain mask, twice as many as positive patches. This procedure resulted in a dataset of ~320k patches for training.

\subsubsection{Network Architecture and Training Procedure}
As depicted in Figure \ref{fig:candDetArch}, we used a seven layers CNN that consisted of four convolutional layers that have 20, 40, 80 and 110 filters of size 7$\times$7, 5$\times$5, 3$\times$3, 3$\times$3 respectively. We applied only one pooling layer of size 2$\times$2 with a stride of 2 after the first convolutional layer since pooling is known to result in a shift-invariance property \citep{scherer2010evaluation}, which is not desired in segmentation tasks. Then we applied three layers of fully connected neurons of size 300, 200 and 2. Finally, the resulting responses were turned into likelihood values using a softmax classifier. We also used batch-normalization \cite{ioffe2015batch} to accelerate the convergence by reducing the internal covariate shift. 

For training the network, we used the stochastic gradient descent algorithm \citep{bottou2010large} with the Adam update rule \citep{kingma2014adam}, mini-batch size of 128 and a categorical cross-entropy loss function. The non-linearity applied to neurons was a rectified linear unit (RELU) to prevent the vanishing gradient problem \citep{maas2013rectifier}. We initialized the weights with the He method \cite{he2015delving}, where the weights are randomly drawn from a $(0, \sqrt{\frac{2}{{fan}_{{\scriptsize in}}}})$ Gaussian distribution. Since CNNs are complex architectures, they are prone to overfit the data very early. Therefore in addition to the batch normalization, we used drop-out \citep{srivastava2014dropout} with 0.3 probability on all fully connected layers as well as $L_2$ regularization with $\lambda_{2}$=0.0001. We used an early stopping policy by monitoring validation performance and picked the best model with the highest accuracy on the validation set.
\begin{figure}[t]
\makebox[\linewidth][c]
{
	\begin{subfigure}[b]{.16\textwidth}
	\centering
	\captionsetup{justification=centering}
	{
	\includegraphics[height=3.9cm]{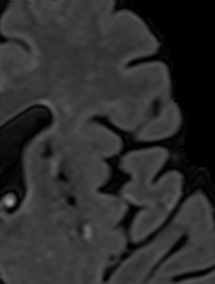} \caption{Original FLAIR image} \label{fig:ROCr1}
	}
	\end{subfigure}
	\begin{subfigure}[b]{.16\textwidth}
	\centering
	\captionsetup{justification=centering}
	{
	\includegraphics[height=3.9cm]{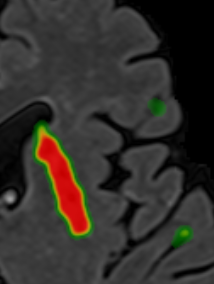} \caption{Candidate segmentation} \label{fig:dice_r1}
	}
	\end{subfigure}
	\begin{subfigure}[b]{.16\textwidth}
	\centering
	\captionsetup{justification=centering}
	{
	\includegraphics[height=3.9cm]{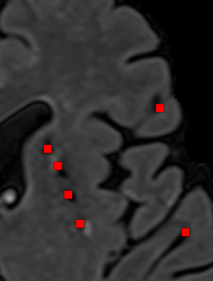} \caption{Candidate extraction} \label{fig:dice_r1}
	}
	\end{subfigure}
}
\caption{An illustrated example on extracting lacune candidates from the (possibly attached) segmentations.}
\label{fig:coarseRecovery}
\end{figure}

\subsubsection{Fully Convolutional Segmentation and Candidate Extraction}
A sliding window patch-based segmentation approach is slow since independently convolving the corresponding patches of neighboring voxels imposes a highly redundant processing. Therefore we utilized a fully convolutional approach for our lacune segmentation. Although the CNN explained in subsection 3.1.2 was trained with patches, we can reformulate the trained fully connected layers into equivalent convolutional filter counterparts \cite{long2015fully}. However, due to the presence of max pooling and convolutional filters the resulting dense prediction is smaller than the original image size. Therefore we used the shift-and-stitch method \cite{long2015fully} to up-sample the dense predictions into a full-size image segmentation. 

A possible coarser segmentation of the candidates might lead to attachment of the segments for two or more close-by candidates. To recover the possibly attached segments into corresponding candidates representative points, we performed a local maxima extraction with a sliding 2D 10$\times$10 window on the likelihoods provided by the CNN (see Figure \ref{fig:coarseRecovery}), followed by a filtering of the local maxima that had a likelihood lower than 0.1. This threshold value was optimized for a compromise between sensitivity and number of extracted candidates on the validation set (0.93 sensitivity with 4.8 candidates per slice on average).

\subsection{False Positive Reduction}
We trained a 3D CNN to classify each detected candidate as either a lacune or a false positive. Contextual information plays an important role for the task at hand as one of the most challenging problems for detection of lacunes, is the differentiation between lacunes and enlarged perivascular spaces. Since perivascular spaces prominently occur in the basal ganglia, location information can be used as a potentially effective discriminative factor. Therefore similar to \cite{ghafoorian2016location}, we employ two mechanisms to provide the network with contextual information: multi-scale analysis and integration of explicit location features into the CNN. Integration of explicit location features has also been recently tried on scoring of coronary calcium in cardiac CT \cite{Wolterink2016123,wolterink2015automatic}, however we add the features to the first fully connected layer that is shown to be more effective \cite{ghafoorian2016location}. These mechanisms will be explained in 3.2.2.

\subsubsection{Sampling}
We captured 3D patches surrounding each candidate at three different scales: 32$\times$32$\times$5, 64$\times$64$\times$5 and 128$\times$128$\times$5 from the FLAIR and T1 modalities, which form the different channels of the input. We down-sample the two larger scale patches to correspond in size with the smaller scale (32$\times$32$\times$5). This is motivated by the main aim of the larger scale patches to provide general contextual information and not the details, which is supplied by smaller scale patch.\\
We used all the lacunes as positive samples and augmented them with cropping all possible 32$\times$32 patches from a larger 42$\times$42 neighborhood and also by horizontally flipping the patches. This yielded an augmentation factor of 11$\times$11$\times$2=242. We randomly picked an equal number of negative samples from non-lacune candidates. To prevent information leakage from the augmentation operation, we applied random cropping for negative samples as well. Otherwise the network could have learned that patches, for which the lacune-like candidate is not located at the center are more likely to be positive. The created input patches were normalized and zero-centered. This sampling process resulted in datasets of ~385k and ~35k samples for training and validation purposes respectively.

\begin{figure*}[ht]
\centering
\centerline
{
	\includegraphics[width=6in]{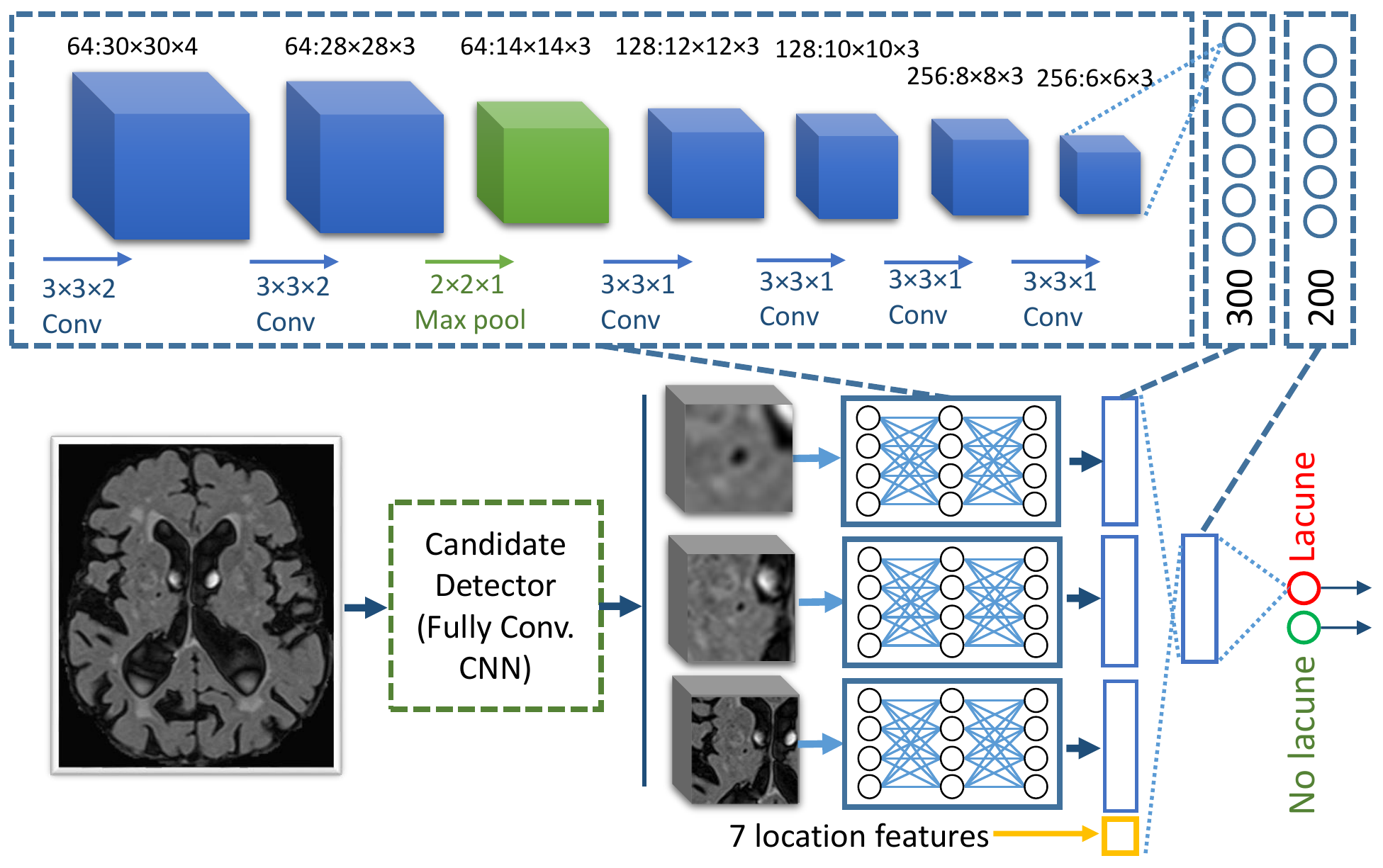}
}
\caption{3D multi-scale location-aware CNN architecture for false positive reduction.}
\label{fig:fprNet}
\end{figure*}

\subsubsection{Network Architecture and Training Procedure}
Referring to Figure \ref{fig:fprNet}, we utilized a late fusion architecture to process the multi-scale patches. Each of the three different scales streamed into stacks of 6 convolutional layers with weight sharing among the streams. Each stack of 6 convolutional layers consisted of 64, 64, 128, 128, 256, 256 filters of size 3$\times$3$\times$2, 3$\times$3$\times$2, 3$\times$3$\times$1, 3$\times$3$\times$1, 3$\times$3$\times$1, 3$\times$3$\times$1 respectively. We applied a single 2$\times$2$\times$1 pooling layer after the second convolutional layer.\\
The resulting feature maps were compressed with three separate fully connected layers of 300 neurons and were concatenated. At this stage, we embedded seven explicit location features to form a feature vector of size 907, which represents a local appearance of the candidate at different scales, together with information about where the candidate is located. The seven integrated features describe for each candidate the $x$, $y$ and $z$ coordinates of the corresponding location in the atlas space, and its distances to several brain landmarks: the right and the left ventricles, the cortex and the midsagittal brain surface. Then the resulting 907 neurons were fully connected to two more fully connected layers with 200 and 2 neurons. The resulting activations were finally fed into a softmax classifier. The activations of all the layers were batch-normalized.

The details of the training procedure were as follows: stochastic gradient descend with Adam update and mini-batch size of 128, RELU activation units with the He weight initialization, dropout rate of 0.5 on fully connected layers and $L_2$ regularization with $\lambda_2$=2e-5, a decaying learning rate with an initial value of 5e-4 and a decay factor of 2 applied at the times that the training accuracy dropped, training for 40 epochs, and selecting the model that acquired the best accuracy on the validation set.

\subsubsection{Test-time Augmentation}
It has been reported that applying a set of augmentations at the test time and aggregating the predictions over the different variants might be beneficial \cite{sato2015apac}. Motivated by this, we also performed test-time augmentation by means of cropping and flipping the patches (as explained in subsection 3.2.1) and then averaged over the predictions for the resulting 242 variants, per sample.

\subsection {Observer Study}
Since an important ultimate goal for the computer-aided diagnosis field is to establish automated methods that perform similar to or exceed experienced human observers, we conducted an observer study, where four trained observers also rated the test set and we compared the performance of the CAD system with the four trained observers. The training procedure was as follows: The observers had a first session on definition of the lacunes, their appearances on different modalities (FLAIR and T1), similar looking other structures such as perivascular spaces and their discriminating features, following the conventions defined in the established standards in SVD research \cite{wardlaw2013neuroimaging}. Then each observer separately rated 20 randomly selected subjects from the training set. In a subsequent consensus meeting, the observers discussed the lacunes they had detected/missed on the mentioned set of images. After the training procedure, each observer independently marked the lacunes by selecting a single representative point for the lacunes appearances on each slice.

\begin{table*}[t]
\centering
\caption{Number of detected lacunes on different definitions of observers agreements and the corresponding sensitivity of the candidate detector on each set. The last four columns represent the reference standards that are formed by excluding each observer and performing majority vote over the remaining observers. The candidate detector detects 4.6 candidates per slice (213 per scan) on average.}
\label{tab:lacunesNum}
\begin{tabular}{@{}lcccccc@{}}
\toprule
\multirow{2}{*}{Measure\textbackslash Reference standard}   & \multicolumn{1}{c}{\multirow{2}{*}{\begin{tabular}[c]{@{}c@{}}At least\\ 2 out of 4\end{tabular}}} & \multirow{2}{*}{\begin{tabular}[c]{@{}l@{}}At least\\ 3 out of 4\end{tabular}} & \multicolumn{4}{c}{\begin{tabular}[c]{@{}c@{}}At least \\ 2 out of 3 excluding\end{tabular}} \\ \cmidrule(l){4-7} 
                               & \multicolumn{1}{c}{}                                                                               &                                                                                & Obs.1                  & Obs.2                  & Obs.3                  & Obs.4                 \\ \midrule
Number of detected lacunes     & 92                                                                                                 & 38                                                                             & 76                    & 81                    & 51                    & 52                   \\
Candidate detector sensitivity & 0.97                                                                                              & 1                                                                              & 0.97                 & 0.98                 & 0.98                 & 0.98                \\ \bottomrule
\end{tabular}
\end{table*}

\subsection {Experimental Setup}
\subsubsection {FROC}
We performed a free-response operating characteristic (FROC) analysis in order to evaluate the performance of the proposed CAD system to compare it to the trained human observers. To be more specific, for comparing the CAD system to the $i$-th observer, we took the observer $i$ out, and formed an evaluation reference standard from the remaining three observers. We used majority voting to form the reference standard, meaning that we considered an annotation as a lacune if at least 2 out of the 3 remaining observers agreed with that. For both CAD and the $i$-th observer to compare with, we considered a detection as a true positive, if it was closer than 3mm to a representative lacune marker in the reference standard, otherwise we counted that as a false positive. Wherever appropriate, we provided with the FROC curves, 95\% confidence intervals obtained through bootstrapping with 100 bootstraps. For each bootstrap, a new set of scans was constructed using sampling with replacement.

\subsubsection {Experiments}
In our experiments we first measured results regarding the observer study, including the number of detected lacunes by each observer, the number of lacunes in several agreement-sets, based on different definitions of agreement, and the performance of our candidate detector (average number of produced candidates and sensitivity on each observer agreement set). Then we evaluated and compared the proposed CAD system with the four available trained human observers using FROC analysis, followed by another FROC analysis for a feasibility study, in which we showed to what extent a trained human observer would benefit from our proposed CAD approach, once the CAD detections are exposed to the observer. To be more specific, the markers of the CAD at a certain threshold with a high specificity (0.88 sensitivity and 0.07 false positives per slice), were shown to the observer who was then asked to check the CAD suggestions, followed by a check to add any other lacune that was missing.

Finally, we show the contribution of two of the components of our method, namely our mechanisms to integrate contextual information (the multi-scale analysis and location feature integration) and the test-time augmentation. To numerically show the contribution of the mentioned method components, we summarize the FROC curves with a single score defined as the average sensitivity over operating points with false positives below 0.4 per slice. We perform this analysis for the reference standards formed by agreement of at least either two or three out of the four observers. For these comparisons, we also provide empirical $p$-values computed based on 100 bootstraps.

\begin{figure*}[!t]
	\makebox[\linewidth][c]
	{
		\begin{subfigure}[b]{.45\textwidth}
		\captionsetup{justification=centering,margin=0.1cm}
		\centering
		{\includegraphics[height=6.5cm]{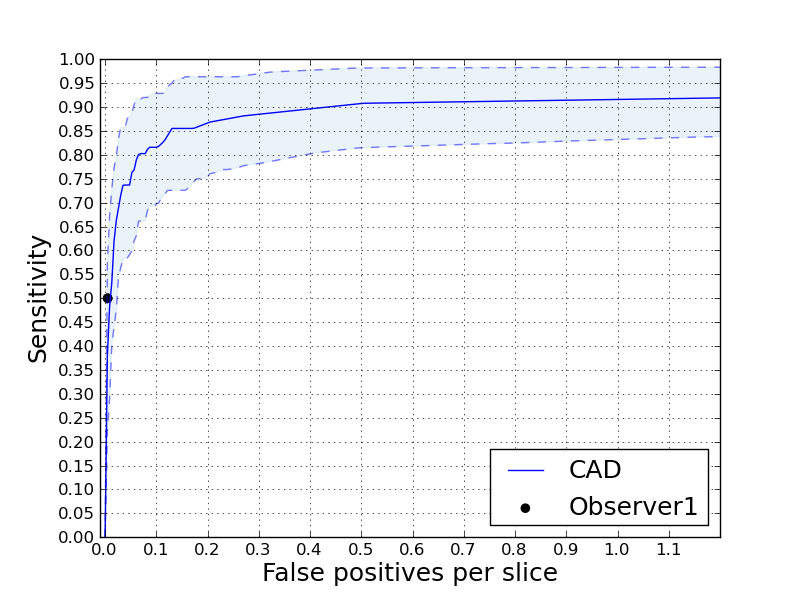}\caption{Comparison with the trained observer 1}\label{fig:FROC_ag2_1}}
		\end{subfigure}
		
		\begin{subfigure}[b]{.45\textwidth}
		\captionsetup{justification=centering,margin=0.1cm}
		\centering
		{\includegraphics[height=6.5cm]{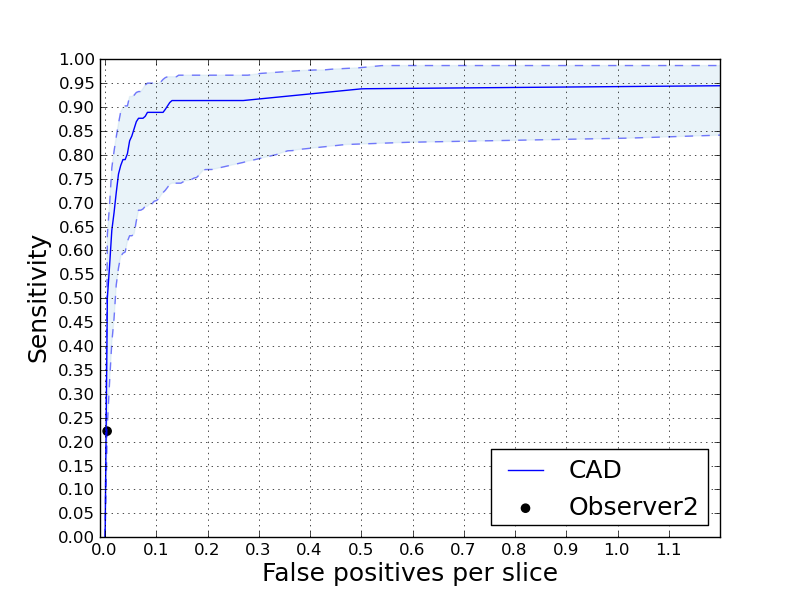}\caption{Comparison with the trained observer 2}\label{fig:FROC_ag2_2}}
		\end{subfigure}
	}
	\makebox[\linewidth][c]
	{
		\begin{subfigure}[b]{.45\textwidth}
		\captionsetup{justification=centering,margin=0.1cm}
		\centering
		{\includegraphics[height=6.5cm]{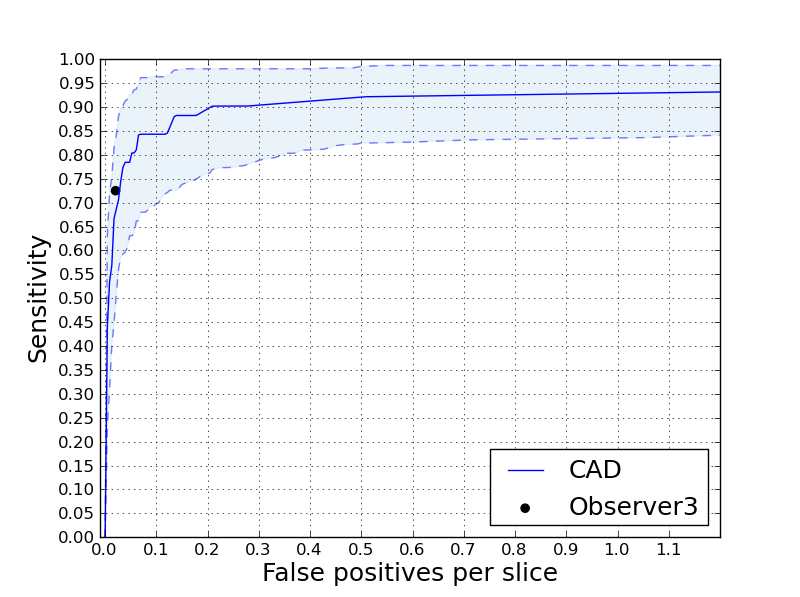}\caption{Comparison with the trained observer 3}\label{fig:FROC_ag2_3}}
		\end{subfigure}
		
		\begin{subfigure}[b]{.45\textwidth}
		\captionsetup{justification=centering,margin=0.1cm}
		\centering
		{\includegraphics[height=6.5cm]{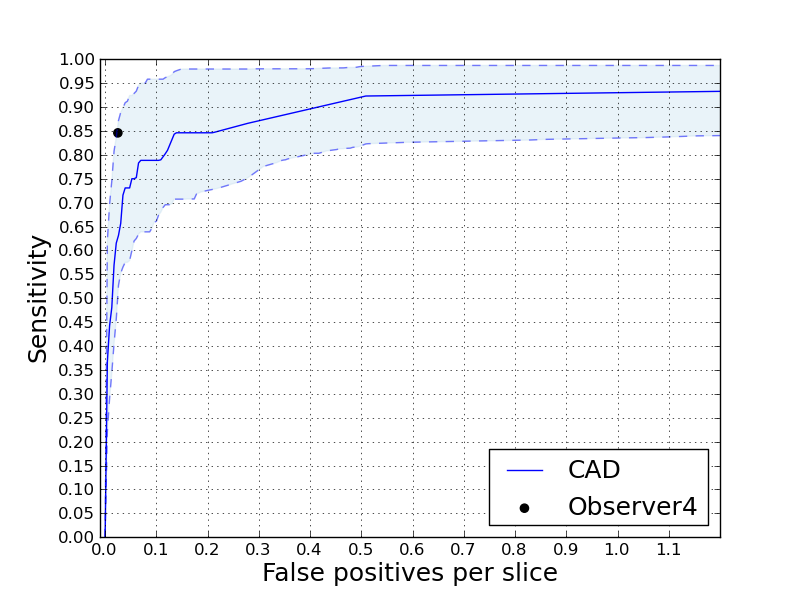}\caption{Comparison with the trained observer 4}\label{fig:FROC_ag2_4}}
		\end{subfigure}
	}	
	\caption{FROC curves comparing the performance of different trained observers with the proposed CAD system. The reference standards for comparing with observer $i$ is formed with the lacunes that at least 2 out of the 3 remaining observers agree on. Shaded area indicates 95\% intervals.}
	\label{fig:FROC_ag2}
\end{figure*}

\section{Results}
It turned out that during the observer study, observers one to four detected 64, 38, 142 and 106 lacune locations respectively. Table \ref{tab:lacunesNum} shows the number of lacunes in agreement between observers, based on different observers agreement definitions, together with the sensitivity of our fully convolutional neural network candidate detector on each agreement set.

Our candidate detector achieves the mentioned sensitivities producing 4.6 candidates per slice (213 per scan) on average. Figure \ref{fig:FROC_ag2} illustrates FROC analyses of the trained observers compared to the corresponding FROC curves for the CAD system, accompanied with 95\% confidence intervals. Figure \ref{fig:withCAD} depicts the difference between the performances of observer 2 with and without observation of CAD marks while detecting the lacunes. 

Figure \ref{fig:threeOutOfFour} provides a more general evaluation of the proposed CAD system using all the four observers to form the reference standard based on majority voting (using lacunes marked by at least 3 out of 4 observers) and also an indication of the contribution of each method components. Table \ref{tab:ingridientBenefit} summarizes this information by reporting $p$-values and scores that represent average sensitivity over operating points with false positives less than 0.4 per slice.

\begin{figure*}[!t]
	\makebox[\linewidth][c]
	{
		\begin{subfigure}[b]{.3\textwidth}
		\captionsetup{justification=centering,margin=0.1cm}
		\centering
		{\includegraphics[height=2.5cm]{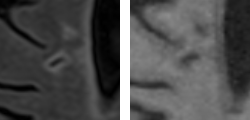}\label{fig:tp3}\caption{ }}
		\end{subfigure}
		
		\begin{subfigure}[b]{.3\textwidth}
		\captionsetup{justification=centering,margin=0.1cm}
		\centering
		{\includegraphics[height=2.5cm]{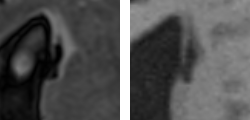}\label{fig:tp2}\caption{ }}
		\end{subfigure}

		\begin{subfigure}[b]{.3\textwidth}
		\captionsetup{justification=centering,margin=0.1cm}
		\centering
		{\includegraphics[height=2.5cm]{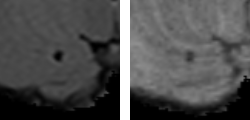}\label{fig:tp1}\caption{ }}
		\end{subfigure}
	}
	\makebox[\linewidth][c]
	{
		\begin{subfigure}[b]{.3\textwidth}
		\captionsetup{justification=centering,margin=0.1cm}
		\centering
		{\includegraphics[height=2.5cm]{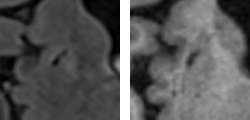}\label{fig:fp1}\caption{ }}
		\end{subfigure}
		
		\begin{subfigure}[b]{.3\textwidth}
		\captionsetup{justification=centering,margin=0.1cm}
		\centering
		{\includegraphics[height=2.5cm]{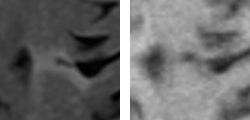}\label{fig:fp3}\caption{ }}
		\end{subfigure}

		\begin{subfigure}[b]{.3\textwidth}
		\captionsetup{justification=centering,margin=0.1cm}
		\centering
		{\includegraphics[height=2.5cm]{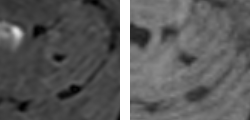}\label{fig:fp2}\caption{ }}
		\end{subfigure}
	}	
	\makebox[\linewidth][c]
	{
		\begin{subfigure}[b]{.3\textwidth}
		\captionsetup{justification=centering,margin=0.1cm}
		\centering
		{\includegraphics[height=2.5cm]{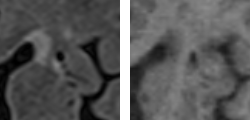}\label{fig:fn1}\caption{ }}
		\end{subfigure}
		
		\begin{subfigure}[b]{.3\textwidth}
		\captionsetup{justification=centering,margin=0.1cm}
		\centering
		{\includegraphics[height=2.5cm]{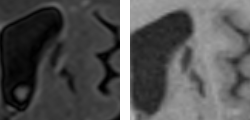}\label{fig:fn2}\caption{ }}
		\end{subfigure}

		\begin{subfigure}[b]{.3\textwidth}
		\captionsetup{justification=centering,margin=0.1cm}
		\centering
		{\includegraphics[height=2.5cm]{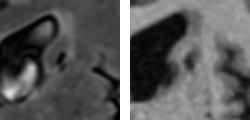}\label{fig:fn3}\caption{ }}
		\end{subfigure}
	}	
	\caption{FLAIR (left) and T1 (right) crops for sample cases of true positives ((a)-(c)), false positives ((d)-(f)) and false negatives ((g)-(i)), with the reference standard formed as the majority of the four observers (at least three out of four), and a threshold of 0.6 (0.7 sensitivity and 0.02 false positives per slice).}
	\label{fig:samples}
\end{figure*}

To provide information about typical true positives, false positives, and false negatives, Figure \ref{fig:samples} illustrates the appearances of the candidates for three sample cases per category on the FLAIR and T1 slices.

\begin{table*}[t]
\centering
\caption{Benefit of context aggregation (multi-scale analysis and location feature integration) and test-time augmentation for the proposed method, analyzed for cases where the reference standard was formed by agreement of at least two or three observers out of four. Scores represent average sensitivity over operating points with false positives less than 0.4 per slice.}
\label{tab:ingridientBenefit}
\begin{tabular}{@{}lcc@{}}
\toprule
Measure \textbackslash Reference standard agreement         & At least 2 out of 4 & At least 3 out of 4 \\ \midrule
Score: proposed CAD                       			& 0.82              & 0.92               \\
Score: no context integration            			& 0.68              & 0.83               \\
$p$-value: with vs. without context integration 	& \textless0.01   & 0.02     \\
Score: no test augmentation                      	& 0.76          	& 0.89               \\
$p$-value: with vs. without test augmentation   & 0.03  		    	& 0.06                \\ \bottomrule
\end{tabular}
\end{table*}

\section{Discussion}
\subsection {Two-stage Approach}
In this study, we used a two-stage scheme with two different neural networks for candidate detection and false positive reduction tasks. The two primary motivations for not using a single network for lacune segmentation are the following: First, the used approach is more computationally efficient. Our much simpler candidate detector network first cheaply removes a vast majority of voxels that are unlikely to be a lacune. Subsequently, we apply a more expensive 3D, multi-scale, location-aware network only on the considerably reduced candidates space (4.6 per slice on average). Second, capturing enough samples from the more informative, harder negative voxels that resemble lacunes (e.g. perivascular spaces) would not be possible in a single stage, due to the resulting training dataset imbalance issue, which requires us to sample with a low rate from the large negative sample pool.

\subsection {Contribution of Method Ingredients}
Referring to Table \ref{tab:ingridientBenefit}, it turns out that providing more contextual information using multi-scale analysis and integrating explicit location features is significantly improving the performance of the resulting CAD approach. This is likely because the appearance of lacunes varies for different brain anatomical locations (e.g. lacunes in the cerebellum usually do not appear with a surrounding hyperintense rim), and the fact that the other similar looking structures are more prominently occurring in specific locations (e.g. perivascular spaces more often appear in the basal ganglia). Such strategies can be effective not only for this particular task, but also in other biomedical image analysis domains, where the anatomical location of the imaging features matters.

\begin{figure}[t]
\centering
\centerline
{
	\includegraphics[width=3.7in]{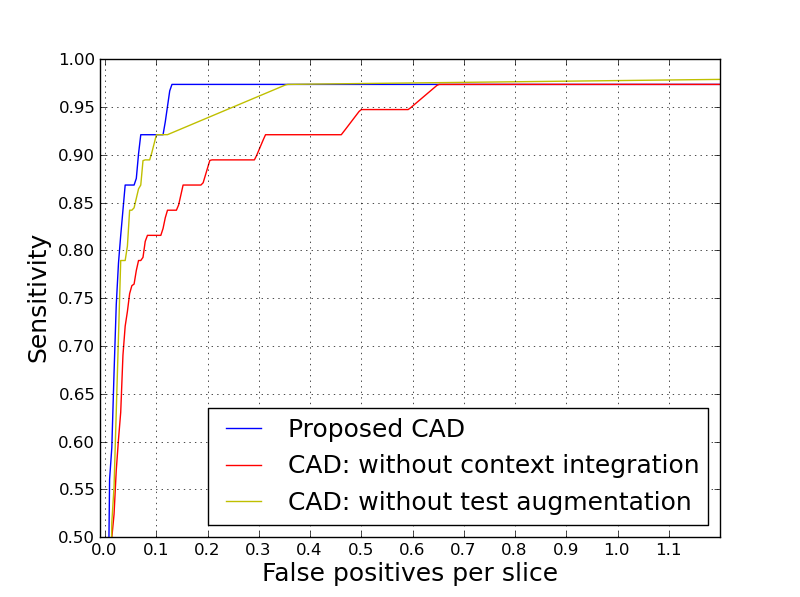}
}
\caption{Contribution of different method components considering agreement of at least 3 out of 4 as the reference standard.}
\label{fig:threeOutOfFour}
\end{figure}

Referring to Table \ref{tab:ingridientBenefit} and Figure \ref{fig:threeOutOfFour}, we observed that test-time augmentation is another effective component. This is likely due to aggregating predictions on an augmented set of pattern representations of a single candidate, reduces the chance that a single pattern in the input space is not well discriminated by the trained neural network.

\subsection {Feasibility Study on Improvement of Human Observers Using CAD}
\begin{figure}[t]
\centering
\centerline
{
	\includegraphics[width=3.7in]{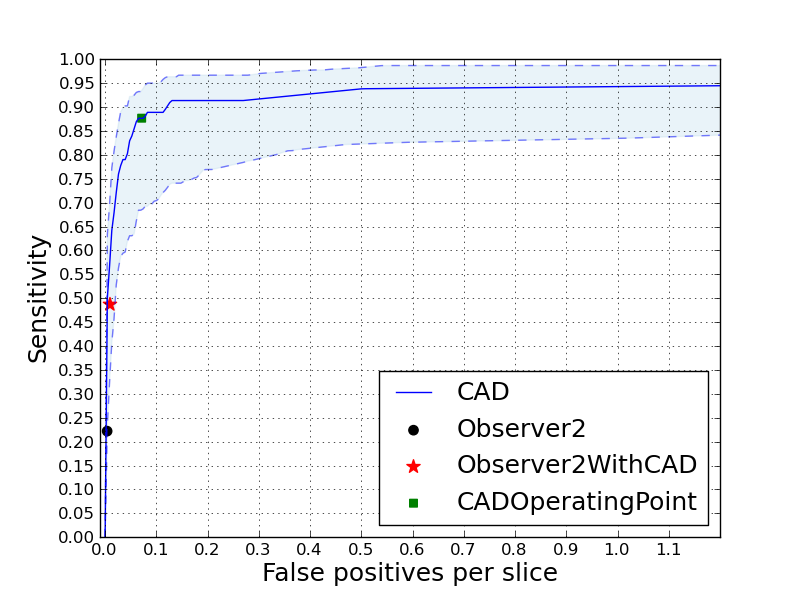}
}
\caption{Improvement of observer 2 once shown the CAD system detections while rating the scans.}
\label{fig:withCAD}a
\end{figure}

Figure \ref{fig:withCAD} shows that a trained human observer can considerably improve once aided by our CAD system. This can be explained by the fact that contrasted by computer systems, humans require a substantial effort for doing an exhaustive search. Therefore showing the markers that the CAD system detects to the human observer, eases the task for the observers and reduces the probability of missing a lacune.

\subsection{Comparison to Other Methods}
As referred to in the introduction section, a number of algorithms with either a rule-based method or supervised learning algorithms with hand-crafted features exist. However, it is not possible to objectively compare the different methodologies on a unified dataset as implementations of none of the methods are publicly available and neither are the datasets these are applied on. Since the majority of the other methods also use FROC analysis, we mention here the reported results on the exclusive datasets just to provide a general idea about the performance of the other methods. Yokoyama et al. \cite{yokoyama2007development} report a sensitivity of 90.1\% with 1.7 false positives per slice on average. The three later methods by Uchiyama et al., using different false positive reduction methods, were all reported to have a sensitivity of 0.968, with 0.76 false positives per slice for the method that used a rule-based and a support vector machine \cite{uchiyama2007computer}, 0.3 false positives for rule-based, neural network and modular classifier \cite{uchiyama2007improvement}, and 0.71 for the eigenspace template matching method \cite{uchiyama2015eigenspace}. At an average false positive of 0.13 per slice, our method detects 97.4\% of the lacunes that the majority of the four observers agree on. We should further emphasize that since the test population's underlying disorder, the MR imaging protocols and the reference standard can influence the results, this does not provide a fair comparison between the different methods. Therefore in our study we chose to compare our automated method to trained human observers that rated the same set of images.

\section{Conclusion}
In this study, we proposed an automated deep learning based method that was able to detect 97.4\% of the lacunes that the majority of the four trained observers agreed on with 0.13 false positives per slice. We showed that integrating contextual information, and test-time augmentation are effective components of this methodology. We also showed in a feasibility study that a trained observer potentially improves when using the presented CAD system.

\section{Acknowledgments}
This work was supported by a VIDI innovational grant from the Netherlands Organisation for Scientific Research (NWO, grant 016.126.351). The authors also would like to acknowledge Inge van Uden, Renate Arntz, Valerie Lohner and Steffen van den Broek for their valuable contributions to this study.

\bibliography{mybibfile}

\end{document}